\pdfoutput=1

\documentclass[11pt]{article}
\usepackage[final]{acl}
\usepackage[compact]{titlesec}         
\titlespacing{\subsection}{0pt}{0pt}{0pt}
\usepackage{times}
\usepackage{latexsym}
\usepackage{algorithm}
\usepackage[noend]{algpseudocode}

\usepackage{multirow} 
\usepackage{caption}
\usepackage[T1]{fontenc}
\usepackage[utf8]{inputenc}

\usepackage{microtype}
\usepackage{inconsolata}
\usepackage{natbib}
\usepackage{float}
\usepackage{graphicx}
\usepackage{placeins}
\usepackage{booktabs}

\usepackage[noabbrev]{cleveref}

\usepackage{tabularx}

\title{Dynamic Jointly Batch Selection for \\
Data Efficient Machine Translation Fine-Tuning}

\author{Mohammad Amin Ghanizadeh and Mohammad Javad Dousti\\
Department of Electrical and Computer Engineering, \\College of Engineering, University of Tehran, Tehran, Iran\\
  \texttt{\{ghanizadeh.amin,mjdousti\}@ut.ac.ir}
}

\begin{document}
\maketitle
\begin{abstract}
Data quality and its effective selection are fundamental to improving the performance of machine translation models, serving as cornerstones for achieving robust and reliable translation systems.
This paper presents a data selection methodology specifically designed for fine-tuning machine translation systems, which leverages the synergy between a learner model and a pre-trained reference model to enhance overall training effectiveness.
By defining a learnability score, our approach systematically evaluates the utility of data points for training, ensuring that only the most relevant and impactful examples contribute to the fine-tuning process.
Furthermore, our method employs a batch selection strategy which considers interdependencies among data points, optimizing the efficiency of the training process while maintaining a focus on data relevance.
Experiments on English $\leftrightarrow$ Persian and several other language pairs using an mBART model fine-tuned on the CCMatrix dataset demonstrate that our method can achieve up to a fivefold improvement in data efficiency compared to an iid baseline. 
Experimental results indicate that our approach improves computational efficiency by $24\%$ when utilizing cached embeddings, as it requires fewer training data points. Additionally, it enhances generalization, resulting in superior translation performance compared to random selection method.

\end{abstract}

\section{Introduction}
\label{sec:introduction}

Machine translation is a fundamental task in natural language processing.
As with any data-driven learning task, the effectiveness of training heavily depends on the quality of the data. \cite{fenza2021data, gupta2021data, chen2021data}
In particular, parallel datasets may contain irrelevant sentence pairs or poorly translated documents, which negatively impact the performance of the final model.

Beyond the quality of data, the state of the learner model itself plays a crucial role in selecting beneficial training data.
For instance, studies have shown that data points associated with high loss on the learner model are typically those the model struggles to learn. \cite{bucher2016hard, kumar2017smart}
Allocating more computational resources to such data points, rather than to those the model has already mastered, can lead to more effective training.

Training can be made more data-efficient by employing selection methods during the training process, such as those based on the loss of data points on the learner model, a pre-trained model, or a combination of both.

We demonstrate that the batch-selection method is more effective than  both the individual sample-selection and random selection method.
More specifically, selecting data points within a batch, where the points are interdependent, is more effective than independently selecting high-scoring data points.
Similar findings have also been reported in previous studies for multimodal learning \cite{evans2024data}. Our experiments focus on 12 different directions, namely,  Persian $\leftrightarrow$ English, German $\leftrightarrow$ English, French $\leftrightarrow$ English, Finnish $\leftrightarrow$ English, Arabic $\leftrightarrow$ English and Hindi $\leftrightarrow$ English.

An mBART model \cite{liu2020multilingual} is used as the learner and a pre-trained LaBSE model \cite{feng2020language} as the reference model.
The pre-trained model is called \textit{reference model}, while the model undergoing fine-tuning is called the \textit{learner model}.

We use features extracted from both the learner model and the pre-trained model for selecting the data during the training.
We employ the learnability score \cite{mindermann2022prioritized} to select data points for fine-tuning.

As demonstrated in our experiments, the use of the learnability score as a selection metric enables the model to generalize more effectively to the data, rather than overfitting. As a result, we achieved up to $5$ times the data efficiency of random selection for English\textrightarrow Persian.

\begin{figure*}[t] 
\centering
\includegraphics[width=0.73\linewidth,height=0.73\textheight,keepaspectratio]{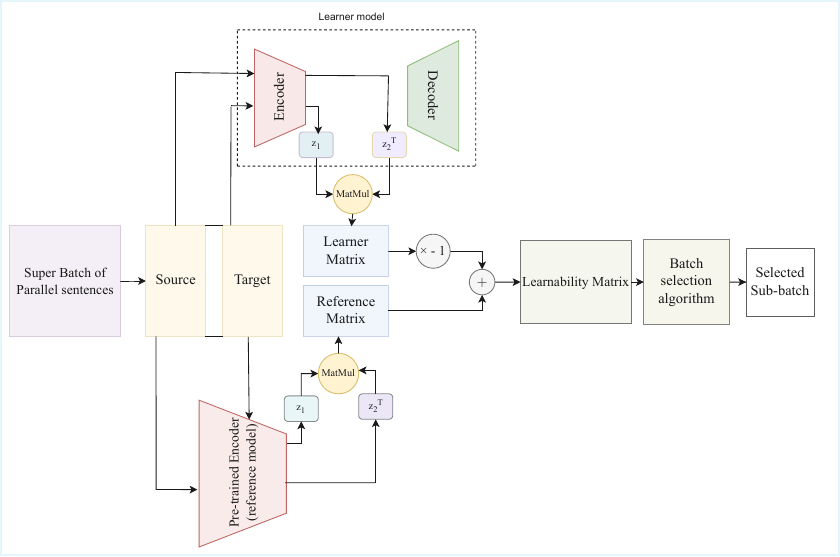}
\caption{Our proposed method diagram for data selection in machine translation}
\vspace{-10pt}
\label{fig:selection-process} 
\end{figure*}

For the remainder of this paper, we refer to training with randomly selected data as iid training. The paper is organized as follows: \Cref{sec:related_work} reviews related work, \Cref{sec:methodology} presents our methodology, \Cref{sec:experiments} details results, and \Cref{sec:conclusion} concludes. \Cref{sec:limitations} discusses limitations. \Cref{sec:appendix_a}, \Cref{sec:appendix_b} and \Cref{sec:appendix_c} contains complementary material.

\section{Related Work}
\vspace{-4pt}
\label{sec:related_work}
\textbf{Offline data selection:}
Traditional methods improve translation and efficiency by selecting parallel data subsets. Studies show that filtering harmful or low-quality data enhances NMT performance \cite{lam2022analyzing, xu2019improving}.

\textbf{Online Data Selection:}
Fixed curation strategies may not adapt to evolving training needs. Online methods dynamically identify challenging examples, improving NMT by varying selected data across training epochs \cite{van2017dynamic}.

\textbf{Hard Negative Mining:}
This technique enhances learning by focusing on difficult negative examples, widely used in computer vision and contrastive learning \cite{bucher2016hard, kumar2017smart, mishchuk2017working, simo2015discriminative, wu2017sampling, xuan2020hard, robinson2020contrastive, 9710301}. However, its application in machine translation remains underexplored.

\textbf{Batch selection.}
Unlike sample selection, batch selection considers inter-data relationships. \citet{evans2024data} proposed an iterative batch selection method using learnability scores in multimodal datasets. Our work extends this concept to machine translation.

\section{Methodology}
\label{sec:methodology}

\subsection{Selection criteria}
Our primary selection criterion is the learnability metric proposed by \citet{mindermann2022prioritized}, consisting of a hard learner score and an easy reference score. The hard learner score is assigned by the learner model, while the easy reference score is assigned by the reference model. We first sample a super-batch of data, ensuring equal selection probability, then choose a sub-batch based on the learnability metric and perform backpropagation.

Effective parallel sentences exhibit closer embeddings in latent space, making similarity between embeddings a key selection factor. A low similarity on the learner model indicates unlearned data points, which should be prioritized. We define the hard learner score as
\begin{equation}
s^{hard}(B, \theta) = - H_{\theta}(B_{src}) H_{\theta}(B_{trg}),
\end{equation}
where $\theta$ denotes learner model parameters, $B$ is the batch and $H_{\theta}(.)$ is the embedding matrix from the learner model. While effective for clean datasets \cite{paul2021deep}, this heuristic can amplify noise in less curated datasets \cite{evans2025bad}.

\begin{figure*}[!t]
\centering
\includegraphics[width=1.0\linewidth,height=1.0\textheight,keepaspectratio]{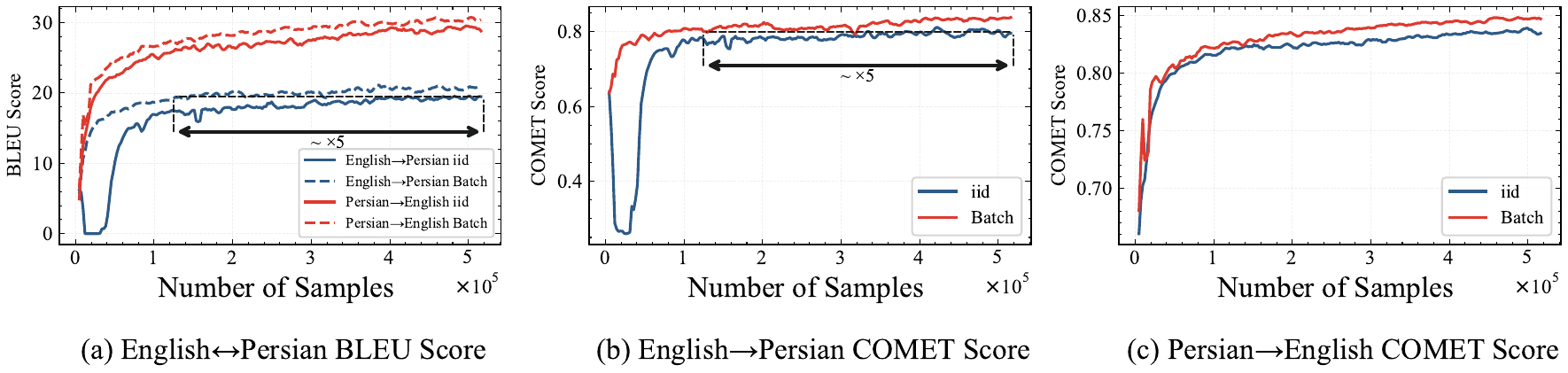}
\caption{Comparison between our approach and independent and identically distributed (iid) training using BLEU and COMET-22 metrics on the filtered dataset.}
\label{fig:comparison-plot}
\end{figure*}

Data points with high similarity on a pre-trained model are typically learnable and high quality \cite{hessel2021clipscore, schuhmann2022laion}. Leveraging this, we filter noisy samples to mitigate overfitting. The easy reference score is defined as
\begin{equation}
s^{easy}(B, \theta^*) = H_{\theta^*}(B_{src}) H_{\theta^*}(B_{trg}),
\end{equation}
where $\theta^*$ represents the reference model parameters. Combining both scores, learnability is defined as
\begin{equation}
s^{learn}(B|\theta,\theta^*) = s^{hard}(B,\theta) + s^{easy}(B, \theta^*).
\label{eq:learning_equation}
\end{equation}
This formulation proritizes unlearned data (high $s^{hard}$) while downweighting noise (low $s^{easy}$).

Similarity is computed as the dot product of sentence embedding from the learner and the reference model, forming matrices. Assuming a super-batch size of $2048$ and embedding dimension of $1024$, this results in $[2048, 1024]$ matrices for both source and target languages. The final similarity matrix, obtained by multiplying these matrices, has a dimension of $[2048, 2048]$. Using this matrix, we compute similarities and derive the learnability matrix via \Cref{eq:learning_equation}.

\begin{algorithm}[t]
\small
\caption{Joint example selection\label{alg:joint_example_selection}}
\begin{algorithmic}[1]
\Require $M_l$ (learnability matrix), $n_{chunks}$, and $filter\_ratio$
\State $C \leftarrow 10^6$ \hspace{1cm} // A large constant
\State $n_{rows} \leftarrow$ \Call{num\_rows}{$M_l$}
\State $n_{draws}\leftarrow\lfloor n_{rows}{\times}(1{-}filter\_ratio){/}n_{chunks}\rfloor$
\State $diag \leftarrow$ \Call{diagonal}{$M_l$}
\State $inds \leftarrow$ \Call{random\_sample}{$diag$, $n_{draws}$}
\For{$z = 1$ \textbf{to} $n_{chunks} - 1$}
    \State $is\_sampled \leftarrow \Call{learnability\_eye}{inds}$
    \State $s_{rows} \leftarrow$ \Call{sum\_rows}{$M_l \times is\_sampled$}
    \State $s_{cols} \leftarrow$ \Call{sum\_columns}{$M_l \times is\_sampled$}
    \State $probs \leftarrow diag + s_{rows} + s_{cols}$
    \State $probs \leftarrow probs - is\_sampled \times C$
    \State $inds' \leftarrow$ \Call{sample\_with\_probs}{$probs$, $n_{draws}$}
    \State $inds \leftarrow$ \Call{concatenate}{$inds$, $inds'$}
\EndFor
\State \Return $inds$
\end{algorithmic}
\end{algorithm}

After computing the learnability matrix, we employ the iterative batch selection algorithm (\Cref{alg:joint_example_selection}) for obtaining the next sub-batch. The algorithm takes the learnability matrix, $n_{chunks}$ (number of data points appended to final mini-batch in each iteration), and a filter ratio as input, outputting selected indices from the super-batch. This approach samples batches that are both learnable and previously unlearned by the model, improving data efficiency compared to individual sample selection, as demonstrated in our experiments.

\section{Experiments}
\label{sec:experiments}

To evaluate our method, we fine-tuned an mBART model on Persian $\leftrightarrow$ English along with German $\leftrightarrow$ English, French $\leftrightarrow$ English, Finnish $\leftrightarrow$ English, Arabic $\leftrightarrow$ English and Hindi $\leftrightarrow$ English subsets of the noisy CCMatrix dataset \cite{nikolova2022filtering}. We considered two settings for Persian $\leftrightarrow$ English: (1) 	\textit{raw dataset fine-tuning}, where mBART was trained on the unprocessed dataset, and (2) \textit{curated dataset fine-tuning}, where CCMatrix was first filtered using LaBSE before applying our method. For other language pairs, we experiment with unfiltered dataset.

\begin{figure*}[!t]
\centering
\includegraphics[width=1.0\linewidth,height=1.0\textheight,keepaspectratio]{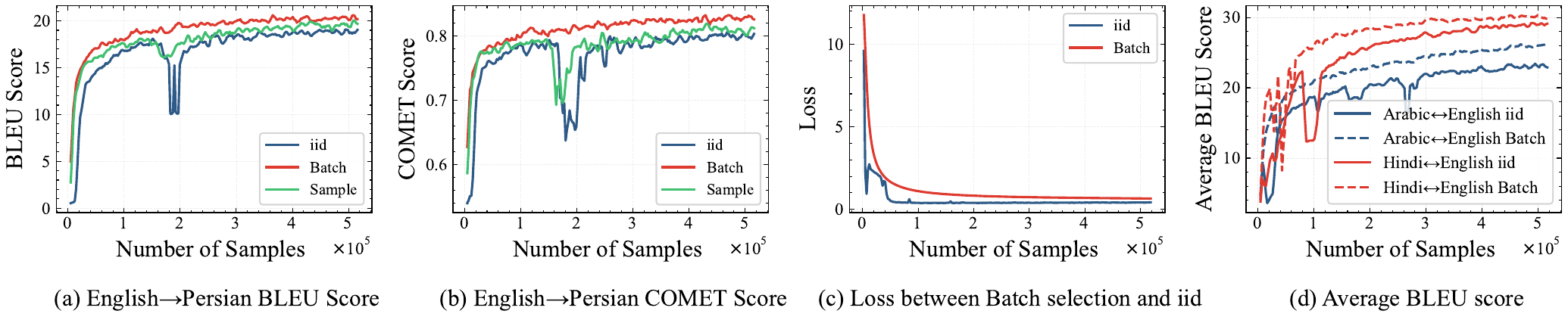}
\caption{(a, b) Comparison of our approach with iid training and individual sample training methods based using BLEU and COMET-22 metrics on the unfiltered dataset.
(c) Batch selection is robust to overfitting on noisy data, especially in early stages of the training.
(d) Comparison of Batch selection and iid on Arabic $\leftrightarrow$ English and Hindi $\leftrightarrow$ English. Each line represents the average of both to and from English directions for each language.}
\label{fig:comparison-nofilter-loss-ar-hi}
\end{figure*}

Our evaluation uses \textsc{FLoRes-200} \cite{guzman-etal-2019-flores}, with all experiments conducted on its test set.
We used a filtering ratio of $0.9$, four chunks, a super-batch size of $4000$, and a sub-batch size of $400$, selecting $400$ samples for updates. Learnability scores of $0.8$ and $0.2$ were used for the reference and learner similarity matrices, respectively.
Smaller super-batches reduced effectiveness, nearing iid performance. 
Final results after training on about $0.5$ million data points are shown in \Cref{tab:final_res}. 
As results demonstrate, batch selection enhances BLEU scores by $1.94$ points for Persian\textrightarrow English direction, whereas it improves English\textrightarrow Persian BLEU score by $1.6$ points.

As shown in \Cref{fig:comparison-plot}, our approach achieves comparable BLEU and COMET-22 scores to that of the iid training, while using approximately five times less data on English\textrightarrow Persian, demonstrating its data efficiency.

\begin{table}[h!]
    \centering
    \resizebox{\columnwidth}{!}{
    \begin{tabular}{@{}lrrrr@{}}
        \toprule
        \multirow{2}{*}{\textbf{Method/Metric}}  & \multicolumn{2}{c}{English\textrightarrow Persian} & \multicolumn{2}{c}{Persian\textrightarrow English} \\ 
       \cmidrule(lr){2-3} \cmidrule(lr){4-5}
         & BLEU & COMET-22 & BLEU & COMET-22 \\ 
        \midrule
        Batch selection  & \textbf{20.86}  & \textbf{0.84} & \textbf{30.32}  & \textbf{0.84}  \\ 
        iid         & 19.26  & 0.78 & 28.38  & 0.83 \\ 
        \bottomrule
    \end{tabular}
    }
    \caption{Final metric for iid and batch selection after training on about $0.5$ million data points for English $\leftrightarrow$ Persian. Results are averaged over two seeds.}
    \vspace{-15pt}
    \label{tab:final_res}
\end{table}

As depicted in Figure~\ref{fig:comparison-nofilter-loss-ar-hi} (c), our batch selection method ensures smoother training loss and improved generalization. By dynamically selecting batches based on learnability, the model avoids overfitting noisy data while maintaining a balanced dataset representation.

We evaluated our approach on unfiltered dataset to test robustness. As shown in Figure~\ref{fig:comparison-nofilter-loss-ar-hi} (a) and (b), joint batch selection is more data-efficient than iid and individual selection, highlighting the benefit of learnability-based batching.

While our method involves more computation than iid training due to extra forward passes, it requires fewer samples to achieve similar performance, resulting in overall efficiency gains when caching reference embeddings (Table~\ref{tab:computation_exp}). Experiments were run on an NVIDIA 3090 GPU, using sub-batch chunks of $32$ samples due to memory limits, though larger sub-batches may improve results further.

\begin{table}[h!]
    \centering
    \small  
    \begin{tabular}{@{}lrr@{}}
        \toprule
        \textbf{Method/Metric}       & \textbf{Samples} & \textbf{Relative FLOPS} \\ 
        \midrule
        Batch selection              & 360,000          & 29.86          \\ 
        Batch selection (cached)     & 360,000          & \textbf{0.76}    \\ 
        iid                          & 1,159,200        & 1.00            \\ 
        \bottomrule
    \end{tabular}
    \caption{Relative floating-point operations with respect to iid training and the number of training samples required to achieve a BLEU score of $21$ on the English\textrightarrow Persian test set.}
    \label{tab:computation_exp}
\end{table}

\subsection{Further experiments in other language pairs}
\begin{figure}[!t]
\centering
\includegraphics[width=\columnwidth,keepaspectratio]{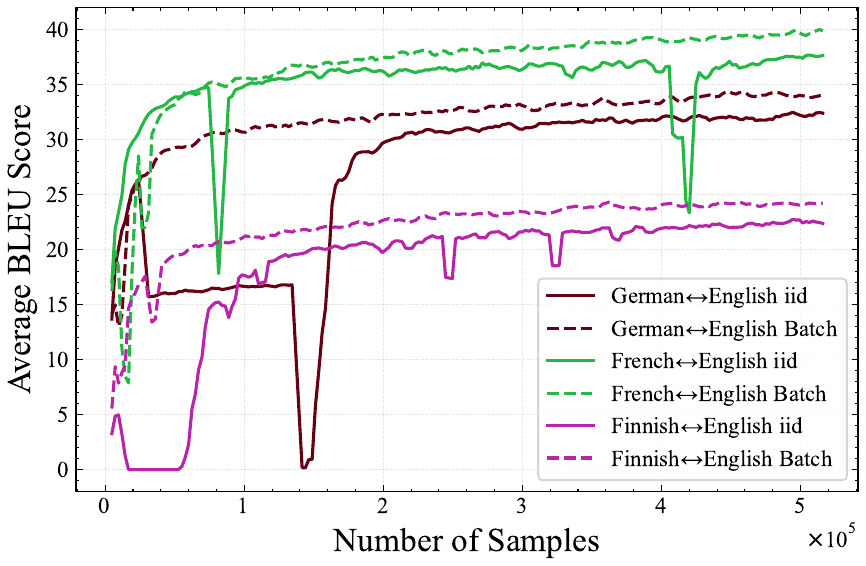}
\caption{Comparison of our approach against iid training on German $\leftrightarrow$ English, French $\leftrightarrow$ English and Finnish $\leftrightarrow$ Englsih. Each line represents the average of both to and from English directions for each language.}
\vspace{-10pt}
\label{fig:eu-blue}
\end{figure}

We further evaluate the effectiveness of our method on additional language pairs and translation directions. As shown in \Cref{fig:eu-blue} and \Cref{fig:comparison-nofilter-loss-ar-hi}~(d), the results demonstrate the robustness of our approach across different languages.

Further experiments are presented in \Cref{sec:appendix_c}.

\section{Conclusion}
\label{sec:conclusion}

We propose a novel online data selection method to improve machine translation fine-tuning. Using a learnability-based batch selection algorithm, our approach identifies data points that are informative yet not fully learned, enhancing training efficiency. Fine-tuning an mBART model on multiple language pairs, we observe improved performance over iid and individual selection strategies.

Our method shows greater resistance to overfitting and more stable loss trends, particularly in early training. By focusing on optimal learning samples, it boosts data and computational efficiency while ensuring stable parameter updates. This demonstrates the value of data selection in low-resource or noisy settings.

\section{Limitations}
\label{sec:limitations}

A key limitation of any data selection method, including ours, is the additional computational overhead required to calculate the utility of individual data points. Our method requires greater computational resources compared to iid when training the model on an equivalent number of data points, particularly when embeddings are not cached. However, the key advantage of our approach lies in its data efficiency; it enables the learner model to achieve comparable performance with fewer data points than the iid training.

Nonetheless, our method may not be optimal in scenarios where a fixed, small, and carefully curated dataset is available. In such cases, iid training could be a more practical choice, as it eliminates the need for utility calculations and avoids the associated computational costs. This trade-off highlights the context-dependent applicability of our method, emphasizing its strengths in situations where data efficiency outweighs computational concerns. 

\section{Acknowledgements}
The ChatGPT-4o Mini model was utilized exclusively for editing purposes in this study.


\bibliography{custom}

\clearpage 

\appendix

\section{Appendix A: Using smaller models as reference model}
\label{sec:appendix_a}

\begin{figure}[t]
\centering
\includegraphics[width=0.9\linewidth,height=0.9\textheight,keepaspectratio]{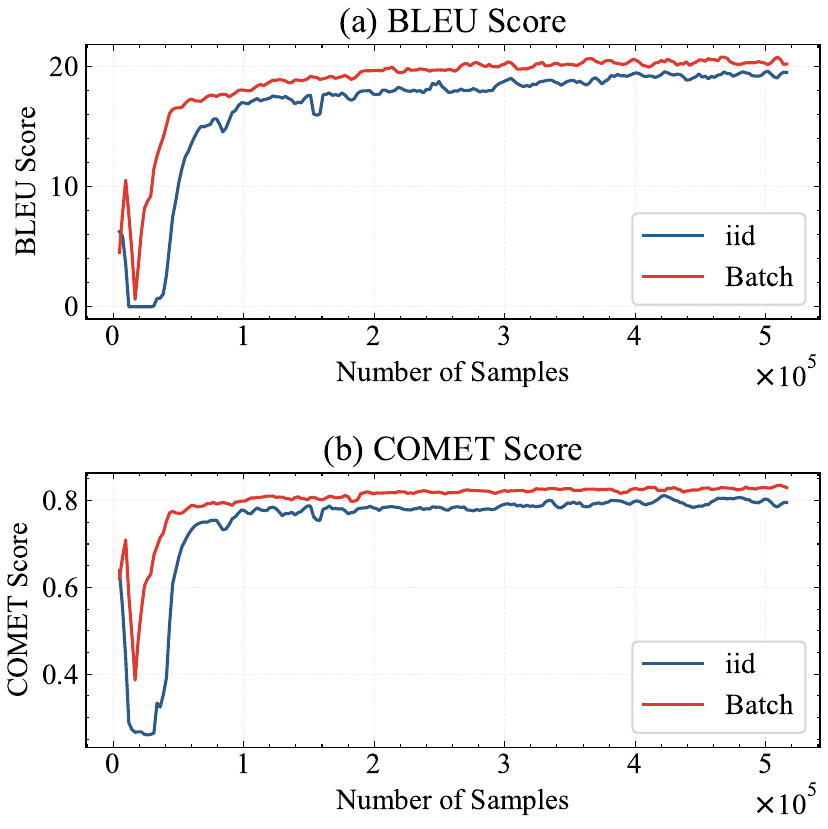}
\caption{We utilize a smaller model as a reference model, apply quantization to it, and demonstrate superior performance compared to iid.}
\label{fig:comparison-nofilter-distiluse}
\end{figure}

To explore computational efficiency, we replaced LaBSE with Distiluse \cite{reimers-2019-sentence-bert} as the reference model. Although Distiluse is significantly smaller, it remained effective for data selection, as shown in Figure~\ref{fig:comparison-nofilter-distiluse}. Furthermore, we applied 4-bit quantization to this model to reduce inference resource requirements. These modifications enabled us to maintain performance while significantly lowering the computational overhead.

This experiment demonstrates that small models are capable of effectively selecting data points for training larger models, as shown in \citet{mekala2024smaller}. This finding highlights the potential of lightweight models in reducing computational costs while maintaining the quality of data selection.

Although smaller models exhibit slight instability at the beginning of training, this issue may be mitigated by adjusting the weights assigned to the learner and reference matrices.

\section{Appendix B: Examining learner and reference scores}
\label{sec:appendix_b}

\begin{figure}[t]
\centering
\includegraphics[width=0.9\linewidth,height=0.9\textheight,keepaspectratio]{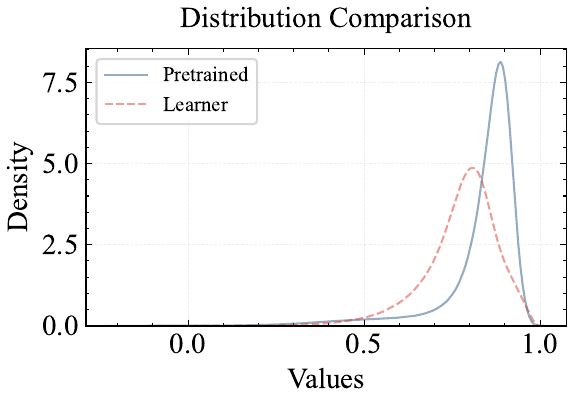}
\caption{Distribution of dot products between the embeddings of source and target sentences.}
\label{fig:dist-plot}
\end{figure}

As stated in the earlier sections, we use dot products between embeddings of the source and target languages as a measure of similarity, where values range between $-1$ and $1$. These scores are then utilized for data selection. For instance, suppose a parallel sentence receives a score of $-1$ from the learner model. According to \Cref{sec:methodology}, we multiply this value by $-1$, yielding a score of $1$. This implies that such a sentence is assigned high priority, despite having an opposite meaning to its counterpart. This scenario could arise if the dataset contained a significant number of parallel sentences with reversed meanings. However, in our case, an analysis of the score distribution demonstrates that this is not the case. Specifically, by measuring and plotting the distribution of dot product values, we observe that very few data points fall below $0$, while the majority of dot product values exceed $0.8$ for both models, as illustrated in \Cref{fig:dist-plot}.

Furthermore, as depicted in \Cref{fig:dist-plot}, the distribution of dot product values for the learner model exhibits a lower mean and higher variance compared to the reference model. This suggests that the learner model remains weaker in its ability to generate aligned embeddings. Ideally, a perfect dataset, when evaluated with a perfect model, would produce a sharp peak at 1, representing an impulse function, indicating that all parallel sentences align perfectly.

\section{Appendix C: Experiment details}
\label{sec:appendix_c}
In this section, we present a comprehensive analysis of the experimental results obtained using our proposed method. We provide a detailed comparison of the performance across various language pairs to highlight the effectiveness and robustness of our approach in multilingual settings.

\begin{figure*}[!t]
\centering
\includegraphics[width=0.8\linewidth,height=0.8\textheight,keepaspectratio]{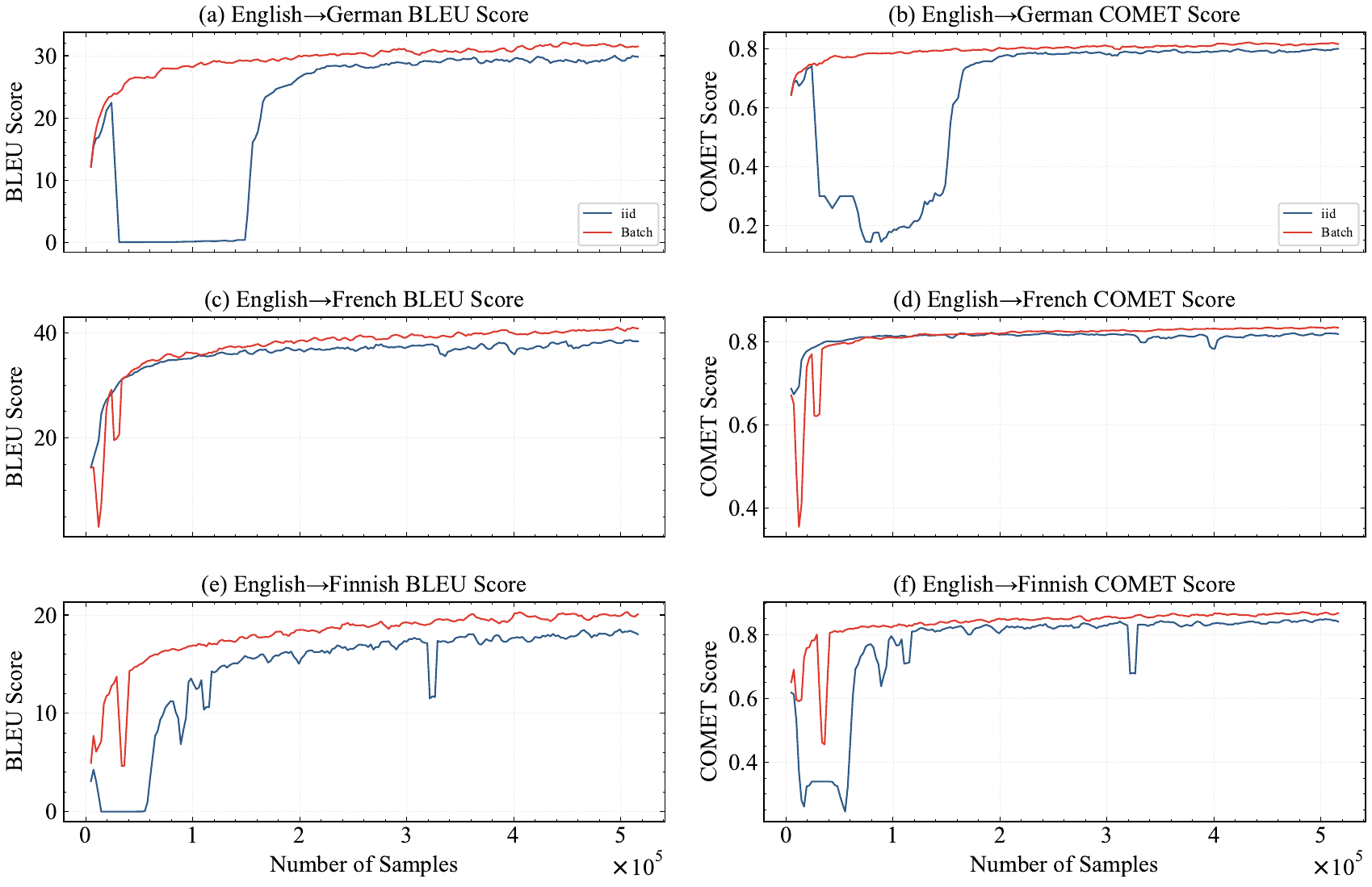}
\caption{Comparison of our approach with iid training on English\textrightarrow German, English\textrightarrow French and English\textrightarrow Finnish.}
\label{fig:comparison-en-to-de-fr-fi}
\end{figure*}

~\Cref{fig:comparison-en-to-de-fr-fi} illustrates the outcomes for translation tasks from English to German, French, and Finnish. We include BLEU score and COMET score to provide a clear view of the model’s strengths.

\begin{figure*}[!t]
\centering
\includegraphics[width=0.8\linewidth,height=0.8\textheight,keepaspectratio]{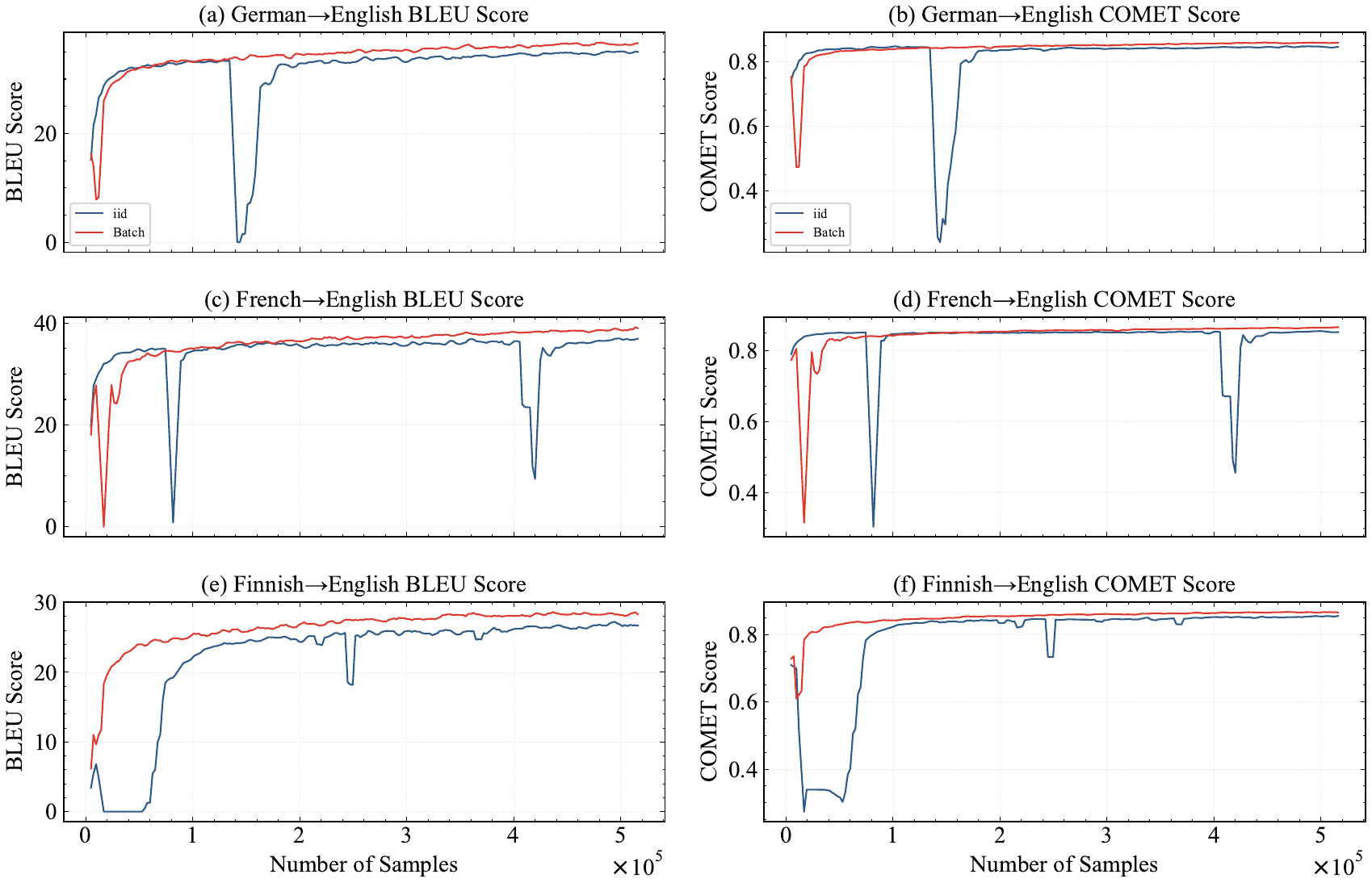}
\caption{Comparison of our approach with iid training on German\textrightarrow English,French\textrightarrow English and Finnish\textrightarrow English.}
\label{fig:comparison-de-fr-fi-to-en}
\end{figure*}

On the other hand, ~\Cref{fig:comparison-de-fr-fi-to-en} reports the same set of metrics, but this time for the reverse direction $xx$\textrightarrow English. This comparison is particularly important for understanding whether the model exhibits any directional bias or asymmetry in translation quality. Notably, the performance in this direction provides insights into the model’s ability to decode diverse linguistic structures back into English.

\begin{figure*}[!t]
\centering
\includegraphics[width=0.8\linewidth,height=0.8\textheight,keepaspectratio]{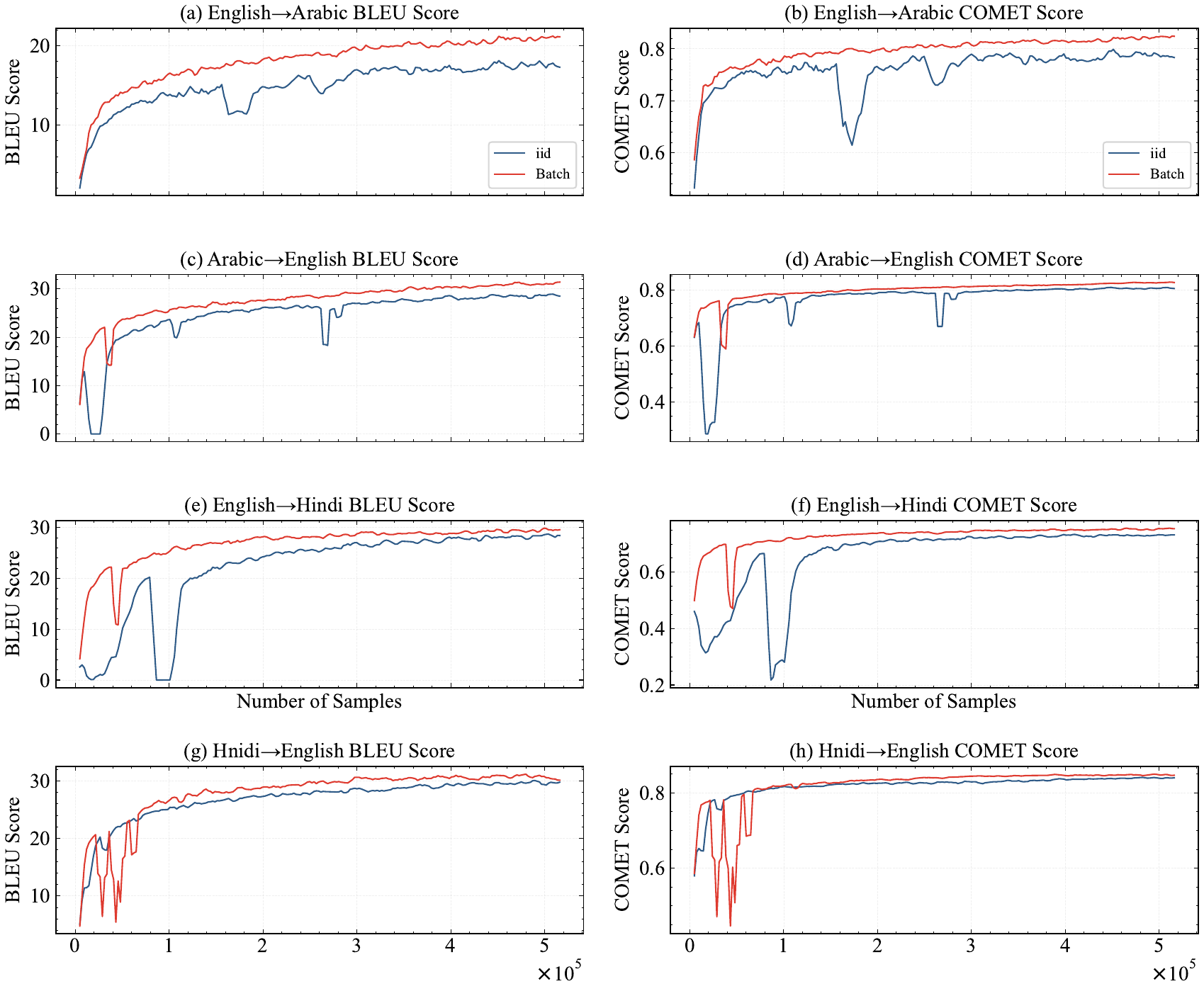}
\caption{Comparison of our approach with iid training on Arabic and Hindi.}
\label{fig:comparison-hi-ar-en}
\end{figure*}

Furthermore, ~\Cref{fig:comparison-hi-ar-en} presents our experimental findings for low-resource languages, specifically Arabic and Hindi. For these languages, we evaluate the model's performance in both translation directions—into and out of English. This helps us assess the model’s generalization capability on typologically distinct languages with limited training data.

Finally, ~\Cref{tab:final_res_hi_ar_en} summarizes the final results of the model for the Arabic and Hindi translation task after the completion of training.

\begin{table*}[h!] 
    \centering
    \begin{tabular}{@{}p{5cm}p{5cm}rr@{}} 
        \toprule
        \textbf{Language} & \textbf{Method} & \textbf{BLEU} & \textbf{COMET-22} \\ 
        \midrule
        \multirow{2}{*}{English\textrightarrow Arabic} & Batch selection & \textbf{21.02} & \textbf{0.82} \\ 
                                          & iid            & 17.43         & 0.78          \\ 
        \midrule

        \multirow{2}{*}{Arabic\textrightarrow English} & Batch selection & \textbf{31.34} & \textbf{0.82} \\ 
                                          & iid            & 28.59         & 0.80          \\ 

        \midrule
        \multirow{2}{*}{English\textrightarrow Hindi} & Batch selection & \textbf{29.52} & \textbf{0.75} \\ 
                                          & iid            & 28.53        & 0.73         \\ 

        \midrule
        \multirow{2}{*}{Hindi\textrightarrow English} & Batch selection & \textbf{30.10} & \textbf{0.84} \\ 
                                          & iid            & 29.94         & \textbf{0.84}          \\ 
                                          
    \end{tabular}
    \caption{Final metric for iid and batch selection after training on about 0.5 million data points for Arabic and Hindi.}
    \label{tab:final_res_hi_ar_en}
\end{table*}

\end{document}